\pdfoutput=1

\documentclass[11pt]{article}

\usepackage[final]{acl}
\usepackage[most]{tcolorbox}
\usepackage{times}
\usepackage{latexsym}
\usepackage{amsmath} 
\usepackage{titlefoot}
\usepackage{pifont} 
\usepackage{marvosym}
\usepackage{authblk}
\usepackage[T1]{fontenc}

\usepackage[utf8]{inputenc}

\usepackage{microtype}

\usepackage{inconsolata}

\usepackage{graphicx}

\title{Word Form Matters: LLMs’ Semantic Reconstruction under Typoglycemia}

\author{Chenxi Wang\textsuperscript{\ding{72}} \quad
    Tianle Gu\textsuperscript{\ding{72}} \quad
    Zhongyu Wei\textsuperscript{$^\clubsuit$}\quad
    Lang Gao\textsuperscript{\ding{72}} \quad
    Zirui Song\textsuperscript{\ding{72}} \quad
    Xiuying Chen\textsuperscript{\ding{72}}\textsuperscript{\Letter}
}

\affil{\textsuperscript{\ding{72}}Mohamed bin Zayed University of Artificial Intelligence (MBZUAI) \quad \textsuperscript{$^\clubsuit$}Fudan University} 

\affil{\texttt{\{chenxi.wang, xiuying.chen\}@mbzuai.ac.ae}}

\begin{document}
\maketitle

\begin{abstract}

Human readers can efficiently comprehend scrambled words, a phenomenon known as \textit{Typoglycemia}, primarily by relying on word form; if word form alone is insufficient, they further utilize contextual cues for interpretation.
While advanced large language models (LLMs) exhibit similar abilities, the underlying mechanisms remain unclear. 
To investigate this, we conduct controlled experiments to analyze the roles of word form and contextual information in semantic reconstruction and examine LLM attention patterns.
Specifically, we first propose \textit{SemRecScore}, a reliable metric to quantify the degree of semantic reconstruction, and validate its effectiveness. 
Using this metric, we study how word form and contextual information influence LLMs’ semantic reconstruction ability, identifying \textit{word form as the core factor} in this process.
Furthermore, we analyze \textit{how} LLMs utilize word form and find that they rely on specialized attention heads to extract and process word form information, with this mechanism remaining stable across varying levels of word scrambling.
This distinction between LLMs’ fixed attention patterns primarily focused on word form and human readers’ adaptive strategy in balancing word form and contextual information provides insights into enhancing LLM performance by incorporating human-like, context-aware mechanisms\footnote{\url{https://github.com/Aurora-cx/TypoLLM}.}.
\end{abstract}

\unmarkedfntext{\Letter: Corresponding Author.}

\section{Introduction}

\textit{User: Do you \underline{undretsand} Typoglycemia?}

\noindent \textit{LLM: Yes! Typoglycemia is a phenomenon where people can still read a word or sentence even when the middle letters of the words are scrambled, ...}

\medskip 

Humans exhibit remarkable adaptability in reading, even when the internal character order of words is scrambled, as long as the first and last letters remain intact.  
This phenomenon, \textit{Typoglycemia}, raises a fundamental question:
\noindent\textit{Why do humans and LLMs understand scrambled words?}  

Research shows that humans recognize words primarily through holistic shape matching~\cite{larson2004science}, relying on contextual prediction when shape cues are insufficient~\cite{sabri_typoglycemia}. In extreme scrambling, higher-order regions like the prefrontal cortex enable retrospective reasoning to infer words from context~\cite{rayner2006raeding}.
Similarly, LLMs exhibit robustness to character scrambling. \cite{cao2023unnatural} found they maintain high accuracy despite disrupted tokenization, while \cite{yu2024mindscrambleunveilinglarge} showed strong Typoglycemia task performance. These findings suggest LLMs, like humans, leverage contextual reasoning to reconstruct scrambled words.

However, a key question remains unanswered: \textit{How do LLMs internally process scrambled text? Do they employ mechanisms similar to humans?}
To investigate this, we designed a series of controlled experiments to systematically analyze the effects of word form and contextual information on LLMs’ semantic reconstruction. 
We constructed a standardized dataset by carefully controlling key linguistic variables and evaluated \textit{LLaMA-3.2 (1B, 3B)-Instruct and LLaMA-3.3 (70B)-Instruct}~\cite{dubey2024llama} to analyze their internal mechanisms underlying this capability.

We define the Semantic Reconstruction Score (SemRecScore) as the cosine similarity between the representation of the original word’s token and the representation of the final subword token of the scrambled word at each layer of the LLM. This metric serves as a reliable measure of semantic reconstruction.
Our results show that LLMs progressively recover word meaning across layers. However, as the degree of word form perturbation increases, semantic reconstruction quality gradually declines. At lower perturbation levels, words achieve near-complete semantic alignment with their original form, whereas at higher perturbation levels, only partial alignment is retained. In the final layers of the models, SemRecScores exhibit significant differences across different perturbation levels.
In contrast, even when contextual information is entirely removed, its impact on semantic reconstruction remains minimal.

Given that LLMs primarily rely on word form for semantic reconstruction, we further investigate how attention mechanisms facilitate this process. Our analysis reveals that attention allocation to word form follows a cyclic pattern across layers. Moreover, as word form perturbation increases, LLMs allocate progressively more attention to word form across all layers, suggesting that reconstructing highly scrambled words requires greater computational resources. Additionally, attention is not uniformly distributed; instead, LLMs rely on specific form-sensitive attention heads dedicated to processing word form information. As perturbation severity increases, more of these specialized attention heads are activated, yet their distribution remains stable, indicating a structured approach to leveraging word form information. 

These findings underscore a key distinction between LLMs and humans: while humans adaptively adjust their reliance on word form and contextual cues based on the degree of perturbation, LLMs primarily rely on word form and exhibit a relatively fixed yet structured attention allocation pattern.

Our contributions can be summarized as follows:
(1) We introduce a new perspective on LLM interpretability by investigating their internal mechanisms under character-level perturbations and propose SemRecScore as a reliable metric for quantifying LLMs' semantic reconstruction ability.
(2) Through systematic experiments, we demonstrate that word form is the primary factor in semantic reconstruction, while contextual information has minimal impact, and we reveal the key role of form-sensitive attention heads in this process.
(3) We further uncover a fundamental divergence between LLMs and human cognition, providing insights for improving LLMs' semantic adaptability.

\section{Related Work}

\textbf{Mechanisms Underlying Human Typoglycemia.} 
Humans’ Typoglycemia ability suggests that word recognition does not strictly depend on letter order but involves higher-level cognitive mechanisms. 
Research in cognitive science and psychology suggests that word recognition primarily relies on word form rather than letter-by-letter decoding~\cite{shillcock2000eye}. 

Eye-tracking studies~\cite{johnson2007transposed,white2008eye} show that fluent readers process words holistically rather than focusing on each letter. As long as the first and last letters remain intact, recognition remains stable~\cite{rayner2006raeding,johnson2012importance}.
Beyond word shape, contextual priming further speeds up recognition~\cite{plummer2014influence}. For example, in “The nurse gave the patient a…”, words like 'doctor' or 'medicine' are automatically activated~\cite{sabri_typoglycemia}, helping readers reconstruct scrambled words more efficiently.
These findings suggest that human readers flexibly adapt between word form and context.

\noindent \textbf{LLM-Based Typoglycemia.}
Some recent studies have explored Typoglycemia and its impact on LLM robustness, but their analyses remain at a surface level, lacking a systematic investigation of internal mechanisms.
For instance, \citet{cao2023unnatural} demonstrated that GPT-4 and other advanced LLMs maintain strong language understanding under extreme character perturbations, accurately reconstructing scrambled words. However, their study was limited to input-output comparisons without examining how LLMs internally process scrambled text. Expanding on this, \citet{yu2024mindscrambleunveilinglarge} evaluated multiple LLMs on Typoglycemia tasks and found them remarkably robust. Yet, their focus remained on task-level performance rather than the underlying representations and attention mechanisms driving word recovery. Thus, while these works show that LLMs can handle Typoglycemia, they offer only observational insights. Our study fills this gap by analyzing LLMs’ internal processing under Typoglycemia perturbations.

\noindent \textbf{Robustness to Word Order Perturbations and Contextual Dependencies.}

\citet{pham-etal-2021-order} and \citet{sinha-etal-2021-masked} found that BERT-based models maintain high accuracy in NLU tasks even when word order is shuffled, indicating a reliance on lexical co-occurrence patterns rather than syntactic structures. Similarly, \citet{gupta2021bert} observed that BERT struggles to detect unnatural inputs, often making high-confidence predictions despite severe word order disruption. However, these studies focus on masked language models and do not extend to generative LLMs or character-level perturbations. \citet{zhu2024context} and \citet{hu2024longcontext} found that pretrained models underperform fine-tuned models in complex reasoning and exhibit overconfidence in long-range dependencies, particularly benefiting from content words and N-grams. Additionally, \citet{hackmann2024wordimportance} and \citet{eisenschlos2024winodict} showed that LLMs primarily rely on statistical correlations rather than explicit syntax and face limitations when acquiring new words beyond their training data. However, these studies do not address how LLMs process and reconstruct text under character-level perturbations.
Building on this work, we systematically analyze the ability of generative LLMs to reconstruct semantics under typoglycemia-style character perturbations, investigating their reliance on word forms and contextual cues while revealing structured attention allocation patterns.

\section{Problem Formulation}

To systematically investigate how LLMs deal with Typoglycemia, we define word semantic reconstruction in LLMs as the internal process through which a model gradually recovers the original meaning of a word from an input with a scrambled character order.
We further break down this problem into two key influencing factors: (1) \textbf{Scramble Ratio.}
\textit{Definition}: The extent of character perturbation within a word, ranging from slight reordering of internal characters (excluding the first and last letters) to extreme scrambling.
\textbf{\textit{Research Question}}: Does increasing scrambling gradually degrade semantic reconstruction across LLM layers?
(2) \textbf{Context Integrity.}
\textit{Definition}: The completeness of contextual information available to the LLM for semantic reconstruction, ranging from a full sentence to no context at all.
\textbf{\textit{Research Question}}: Does LLM semantic reconstruction primarily rely on word shape or context? If context integrity decreases, is the model’s semantic recovery ability significantly affected?

To explore LLMs' processing of scrambled words, we apply the following methods to analyze semantic reconstruction.

(1) \textbf{Attention Distribution Analysis.}
\textit{Objective}: To study how LLMs distribute attention between word shape information and contextual information across different layers, and whether there are hierarchical shifts.
\textit{\textbf{Research Questions}}:
Does LLM adjust its attention patterns based on the degree of word scrambling?
Is there a dynamic weighting shift from word shape reliance to contextual reliance as scrambling increases?
(2) \textbf{Role of Specialized Attention Heads.}
\textit{Objective}: To identify whether specific attention heads are specialized for processing word shape or context and analyze their behavioral patterns across different layers.
\textit{\textbf{Research Questions}}:
Under different scrambling degrees, which attention heads remain highly focused on word shape information?
Which attention heads continue to function effectively under extreme scrambling conditions?

\section{Dataset}\label{sec:dataset}
We selected SQuAD~\cite{rajpurkar-etal-2016-squad} as our base dataset for its extensive use in evaluating LLMs' natural language understanding. Its long, well-structured texts enable analysis across linguistic structures while ensuring clear semantics and controlled scrambling, enhancing reproducibility and comparability.

\subsection{Variables Definition}
In this study, we control two key variables, Scramble Ratio (SR) and Context Integrity (CI), to construct an SR × CI matrix experimental design, enabling a systematic investigation of LLMs' semantic reconstruction ability under different linguistic structure variations.

\textbf{Scramble Ratio (SR).}  
SR quantifies the degree to which the internal characters of a word (excluding the first and last letters) have been scrambled. It is defined as:  
\[
SR = \frac{N_{\text{scrambled}}}{N_{\text{candidate}}},
\]
where \(N_{\text{scrambled}}\) is the number of scrambled characters, and \(N_{\text{candidate}}\) is the total number of characters eligible for scrambling within the word. A higher SR indicates a more disrupted word structure. We define five SR levels: 0, 0.25, 0.5, 0.75, and 1.  

\textbf{Context Integrity (CI).}  
CI measures the completeness of contextual information provided for a given word. It is defined as:  
\[
CI = \frac{N_{\text{context\_preserved}}}{N_{\text{total\_context}}},
\]
where \(N_{\text{context\_preserved}}\) is the number of preserved context words, and \(N_{\text{total\_context}}\) is the total number of words in the original sentence. A higher CI indicates a more complete contextual environment. We define five CI levels: 0, 0.25, 0.5, 0.75, and 1.

\subsection{Data Standardization and Control}
To ensure the reliability of the experimental data, we performed standardization on the dataset.

\paragraph{Word Scrambling.}
All target words are at least 10 characters long to avoid short words affecting results. Based on the SR value, we extract, shuffle, and reinsert a continuous substring from the candidate characters, ensuring the modified sequence differs from the original.
This keeps the word shape intact at the beginning and end, making it easier to quantify. To ensure control and reproducibility, we select only words that remain intact after tokenization, avoiding subword splitting.

\paragraph{Context Masking.}
To study LLMs' semantic reconstruction under context loss, we control CI. Unlike BERT, LLaMA models lack a mask token and was not trained with masked language modeling, making direct word removal problematic as it alters sentence length and disrupts syntax.
To maintain sentence structure and prevent tokenization issues, we replace masked words with `\_'. Masked words are selected based on CI values, ensuring systematic coverage. To reduce noise from masking key contextual words, we generate multiple masked datasets using different random seeds. However, results remain stable across seeds, as large-scale sampling offsets individual sample noise.

\paragraph{Data Examples.}
We selected 20,000 samples from the base dataset and, after standardization, obtained 7,556 qualified samples. 
Below is an example with SR = 0.5:  
The original word \textit{relationship} was scrambled into \textit{relatinioshp}, while keeping the first and last letters unchanged.
Below is an example with CI = 0.5:
\begin{tcolorbox}[colback=gray!10, left=1mm, right=1mm, top=1mm, bottom=1mm] 
\small 
\textbf{[Original Sentence]}  
``...During Franco's regime, however, the blaugrana team was granted profit due to its good relatinioshp with the dictator at management level, even giving two awards to him...''
\end{tcolorbox}

\begin{tcolorbox}[colback=gray!10, left=1mm, right=1mm, top=1mm, bottom=1mm] 
\small  
\textbf{[Processed Sentence (SR = 0.5, CI = 0.5)]}  
``...During Franco's regime, \_ the blaugrana \_ \_ \_ profit \_ \_ \_ \_ relatinioshp with the \_ \_ management \_ even \_ \_ \_ to him...''
\end{tcolorbox}

\section{How to Measure the Degree of Semantic Reconstruction}\label{sec:validate}

In this section, we introduce Semantic Reconstruction Score (SemRecScore) to quantify semantic reconstruction across LLM layers. To validate its effectiveness, we propose Negative Correlation Rate (NegCorrRate), which measures how increased SemRecScore aligns with greater consistency in completion probability. 

We evaluate NegCorrRate across three LLMs (LLaMA-3.2(1B, 3B)-Instruct, LLaMA-3.3(70B)-Instruct) and all samples, empirically supporting SemRecScore’s effectiveness.

\subsection{Semantic Reconstruction Score}

We define \textit{SemRecScore} as a metric to quantify the degree of semantic reconstruction for scrambled words across different layers of an LLM. 

Since scrambled words rarely exist in the tokenizer’s vocabulary, they are typically split into multiple subword tokens, dispersing their representation across embeddings. The model must integrate these fragments to reconstruct meaning. Prior studies suggest that in subword tokenization, the last token in a sequence carries the most integrated semantic representation~\cite{meng2022locating,geva2023dissecting,yang2024large},. Based on this, we define SemRecScore by comparing the original word’s token representation with the last token in the scrambled word’s subword sequence.

Formally, let \( x^{(L)}_o \) be the representation of the original word’s token at layer \( L \), and let \( x^{(L)}_s \) be the representation of the last subword token in the scrambled sequence at the same layer. Then, SemRecScore at layer \( L \) is defined as:
\[
\text{SemRecScore}^{(L)} = \frac{x^{(L)}_o \cdot x^{(L)}_s}{\|x^{(L)}_o\| \|x^{(L)}_s\|},
\]
where \( x^{(L)}_o \cdot x^{(L)}_s \) denotes the dot product between the two vectors, and \( \|x^{(L)}_o\| \) and \( \|x^{(L)}_s\| \) represent their respective Euclidean norms.
The resulting cosine similarity ranges from -1 to 1, with higher values indicating stronger semantic alignment.

SemRecScore provides insights into how well an LLM reconstructs the semantics of a word when its subword structure is disrupted. A higher SemRecScore suggests that the model retains the semantic meaning despite scrambling, while a lower score indicates a loss of semantic integrity.

\subsection{Validating SemRecScore}

To evaluate whether SemRecScore effectively captures semantic reconstruction, we analyze its relationship with the consistency of the model’s behavior in generating completions. Specifically, we introduce \textit{Negative Correlation Rate (NegCorrRate)}, a global statistical measure that quantifies whether an increase in SemRecScore corresponds to greater consistency in the model’s completion probability distribution. This section details the formulation of NegCorrRate and its empirical trends.

\begin{figure}[t]
  \includegraphics[width=\columnwidth]{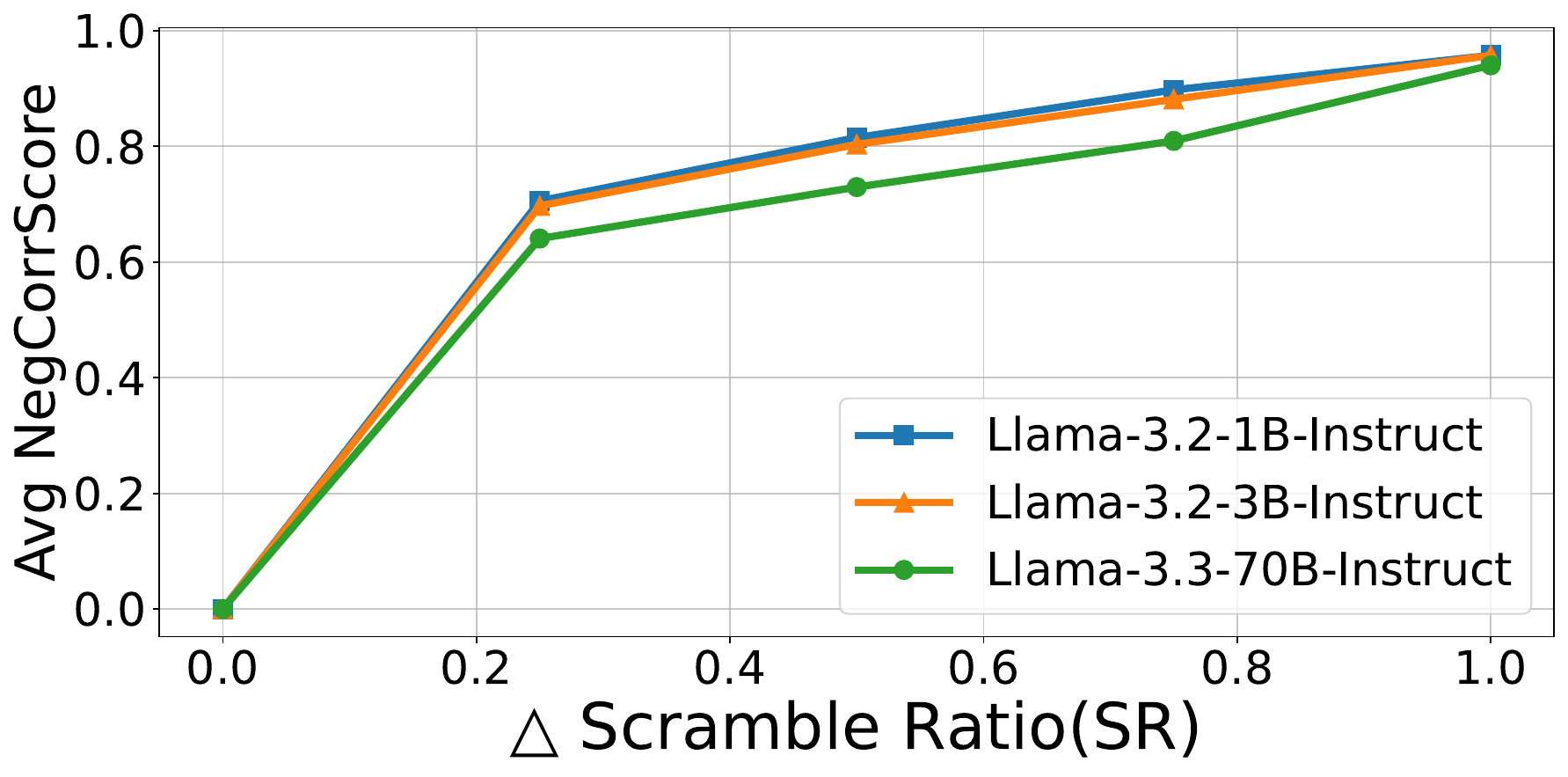}
  \caption{Relationship between $\Delta$SR and Average NegCorrScore across LLaMA models of different scales. The increasing trend of NegCorrScore with $\Delta$SR \textit{validates SemRecScore as a reliable measure of semantic reconstruction}.}

  \label{fig:q1}
\end{figure}

\paragraph{Formulation.}
NegCorrRate is designed to measure the extent to which a higher SemRecScore is associated with a lower KL divergence between the completion probability distributions of prompts containing scrambled and original words. Formally, given a word with multiple Scramble Ratio (SR) levels, we denote the final-layer SemRecScore at SR level \( i \) as \( \text{SRScore}^{(\text{final})}_i \), and the KL divergence between the completion probability distributions of prompts containing the scrambled word at SR level \( i \) and those containing the original word as \( \text{KLdiv}_i \). For any two SR levels \( i \) and \( j \) such that \( j = i - \Delta SR \), where \( \Delta SR \) is a predefined scrambling difference, we compute:
\[
\resizebox{0.48\textwidth}{!}{$
\mathcal{C}_{i,j}=(\text{SRScore}^{(\text{final})}_i - \text{SRScore}^{(\text{final})}_j) \times (\text{KLdiv}_i - \text{KLdiv}_j)
$}.
\]
\(\mathcal{C}_{i,j}\) indicates whether a decrease in final-layer SemRecScore (\(\text{SRScore}^{(\text{final})}_i < \text{SRScore}^{(\text{final})}_j\)) corresponds to an increase in KL divergence (\(\text{KLdiv}_i > \text{KLdiv}_j\)), supporting the expectation that lower semantic reconstruction leads to greater inconsistency in completion behavior.

NegCorrRate is then defined as the proportion of sample pairs where this term is negative:
\[
\text{NegCorrRate} = \frac{1}{|\mathcal{P}|}\textstyle \sum_{(i,j) \in \mathcal{P}} \mathbf{1} (\mathcal{C}_{i,j} < 0),
\]
where \( \mathcal{P} \) is the set of all valid SR level pairs \( (i, j) \) corresponding to a fixed \( \Delta SR \), and \( \mathbf{1}(\cdot) \) is the indicator function that returns 1 when the condition inside holds and 0 otherwise.

\paragraph{Empirical Analysis.}

To understand how the relationship between SemRecScore and model consistency evolves with increasing scrambling severity, we compute the average NegCorrRate across all target words for different values of \( \Delta SR \). The results for three LLaMA models are shown in Figure~\ref{fig:q1}.  

At \( \Delta SR = 0 \), NegCorrRate is 0, as comparing a word with itself introduces no behavioral difference. As \( \Delta SR \) increases, greater scrambling differences lead to larger semantic reconstruction gaps, increasing the number of negatively correlated sample pairs. Consequently, NegCorrRate rises, reinforcing the negative correlation between SemRecScore and KL divergence.  

Figure~\ref{fig:q1} shows a sharp increase in NegCorrRate for small \( \Delta SR \), indicating that even mild scrambling disrupts semantic reconstruction. As \( \Delta SR \) nears 1.0, NegCorrRate approaches 1.0, suggesting a nearly universal negative correlation across samples.
These results provide empirical support that SemRecScore effectively reflects the degree of semantic reconstruction, as its correlation with completion consistency remains robust across different scrambling intensities.

\section{How Word Form and Contextual Information Influence LLMs’ Semantic Reconstruction}\label{sec:exp2}

\begin{figure*}[tb]
    \centering
    \includegraphics[width=1\linewidth]{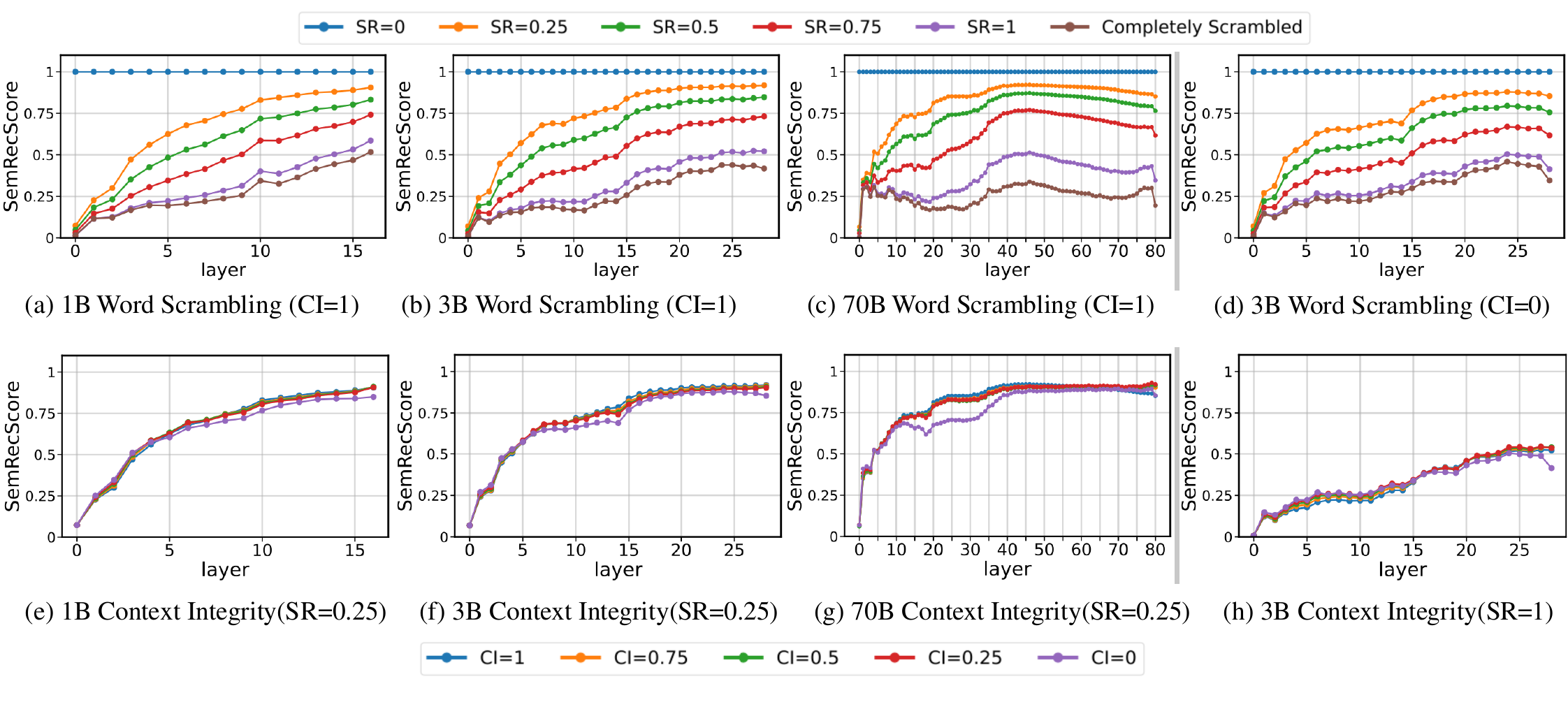}
    \caption{Semantic reconstruction performance across different Scramble Ratios (SR) and Context Integrity (CI) levels. The top row (a-d) presents SemRecScore trends under varying SR values for 1B, 3B, and 70B models. The bottom row (e-h) illustrates SemRecScore evolution for fixed SR values while varying CI. \textit{Across all models, word form plays a dominant role, with context integrity having minimal impact on reconstruction performance.}}
    \label{fig:main}
\end{figure*}

To quantitatively assess the impact of \textit{word form} and \textit{contextual information} on LLMs' semantic reconstruction, we conduct experiments on three instruction-tuned models: LLaMA-3.2(1B, 3B)-Instruct and LLaMA-3.3(70B)-Instruct. We design a \(5 \times 5\) experimental matrix with five levels of Scramble Ratio (SR) and five levels of Context Integrity (CI), yielding 25 distinct settings. Analyzing semantic reconstruction across all layers, we find that \textit{word form plays a dominant role}, while contextual information, though initially expected to be key, has a surprisingly limited effect on LLMs' typoglycemia capabilities (Figure~\ref{fig:main}). The following sections explore the influence of word form first, then contextual information.

\subsection{Impact of Word Form on Semantic Reconstruction}

To assess word form's impact, we analyze Figures~\ref{fig:main}(a)-(c) for 1B, 3B, and 70B models with CI=1.  
Across all models, when SR = 0, no reconstruction is needed, and SemRecScore remains near 1 across layers, confirming token embeddings align with original words. When SR > 0, Layer 0 representations are unrelated to the original forms, regardless of scrambling severity. Reconstruction improves with depth but depends on SR—lower SR values recover more effectively. By the final layer, SR = 0.25 reaches near-perfect reconstruction, while SR = 1 lags by 30\% and only achieves a final SemRecScore of 0.5, indicating incomplete reconstruction. The widening gap between SR levels at deeper layers highlights word form’s critical role in semantic recovery.  
Figure~\ref{fig:main}(c) reveals a 70B model anomaly—unlike 1B and 3B, which show a monotonic increase, highly scrambled words in the 70B model decline in later layers. This suggests larger models reinterpret highly perturbed words as semantically unrelated rather than reconstructing them, revealing a scale-dependent phenomenon where extreme perturbations are disregarded rather than forced into alignment.  
Figure~\ref{fig:main}(d) examines the 3B model with no context (CI=0), showing similar SemRecScore trends as CI=1. Despite context removal, final reconstruction quality aligns with SR-based expectations, indicating minimal contextual impact—words are reconstructed regardless.

These findings demonstrate that word form dominates LLMs’ typoglycemia capabilities, with lower SR aiding recovery, while context plays a minimal role. Additional results for intermediate CI values are in the Appendix~\ref{exp2_app1}.

\subsection{Impact of Contextual Information on Semantic Reconstruction}

To assess the role of contextual information in semantic reconstruction, we analyze Figures~\ref{fig:main}(e), (f), and (g), which correspond to the 1B, 3B, and 70B models under SR=0.25. Notably, we observe that the curves for different CI levels are almost overlapping, indicating that under the same scramble ratio, the completeness of contextual information has minimal impact on semantic reconstruction. The only exception is when CI=0, where a slight drop in reconstruction performance is visible when no context is retained.

Further, Figure~\ref{fig:main}(h) shows the 3B model’s reconstruction trends at SR=1, where all internal characters are scrambled except the first and last. Interestingly, even under this extreme perturbation, the trends for different CI values remain nearly identical, indicating that LLMs do not increase their reliance on contextual information when processing highly scrambled words. This reinforces the observation that contextual integrity does not significantly affect the model’s ability to reconstruct word meaning. However, when comparing Figure~\ref{fig:main}(h) (SR=1) with Figure~\ref{fig:main}(f) (SR=0.25), we observe a clear decline in reconstruction performance, highlighting that word form, rather than context, is the primary determinant of semantic recovery. Additional results for intermediate SR values are in the Appendix~\ref{exp2_app2}.

\section{How LLMs Utilize Word Form Information}

Building on the findings from Section~\ref{sec:exp2}, we have established that word form is the primary factor influencing LLMs’ reconstruction of a word’s original meaning in typoglycemia scenarios. In this section, we further investigate how LLMs utilize word form information.

Specifically, in Section~\ref{sec:attention_allocation}, we analyze the overall attention distribution across different scrambling levels and how it evolves across model layers. In Section~\ref{sec:form_sensitive_heads}, we identify specific attention heads responsible for processing word form, revealing the most fine-grained mechanisms through which LLMs leverage form-based cues.

\begin{figure*}[tb]
    \centering
    \includegraphics[width=1\linewidth]{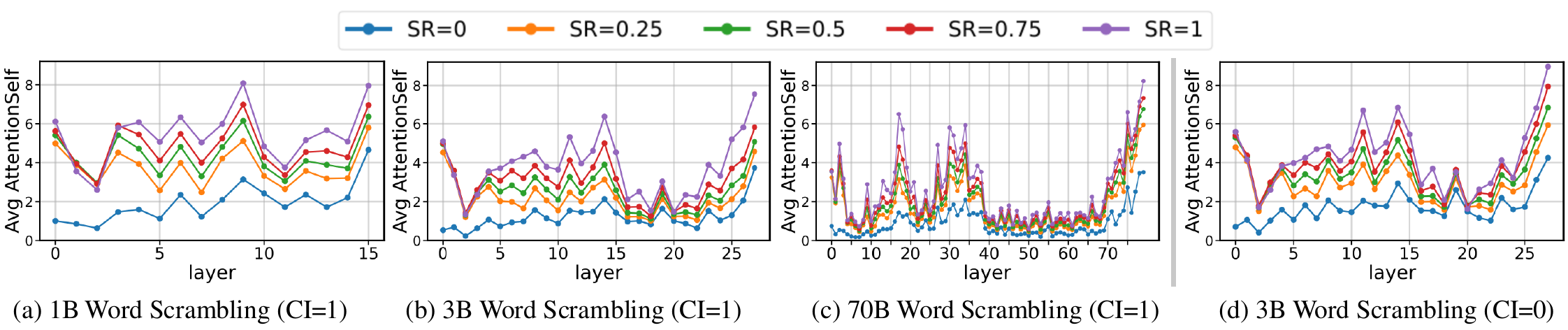}
    \caption{Attention allocation to word form under varying Scramble Ratios (SR).
Subplots (a-c) show AttentionSelf trends for 1B, 3B, and 70B models with full context (CI=1), while (d) presents the 3B model without context (CI=0). Higher SR values consistently elicit stronger attention to word form, and the cyclic attention pattern remains unchanged even without context, \textit{suggesting that LLMs process word form independently of contextual information.}}
    \label{fig:p3_1}
\end{figure*}

\begin{figure*}[htb]
    \centering
    \includegraphics[width=1\linewidth]{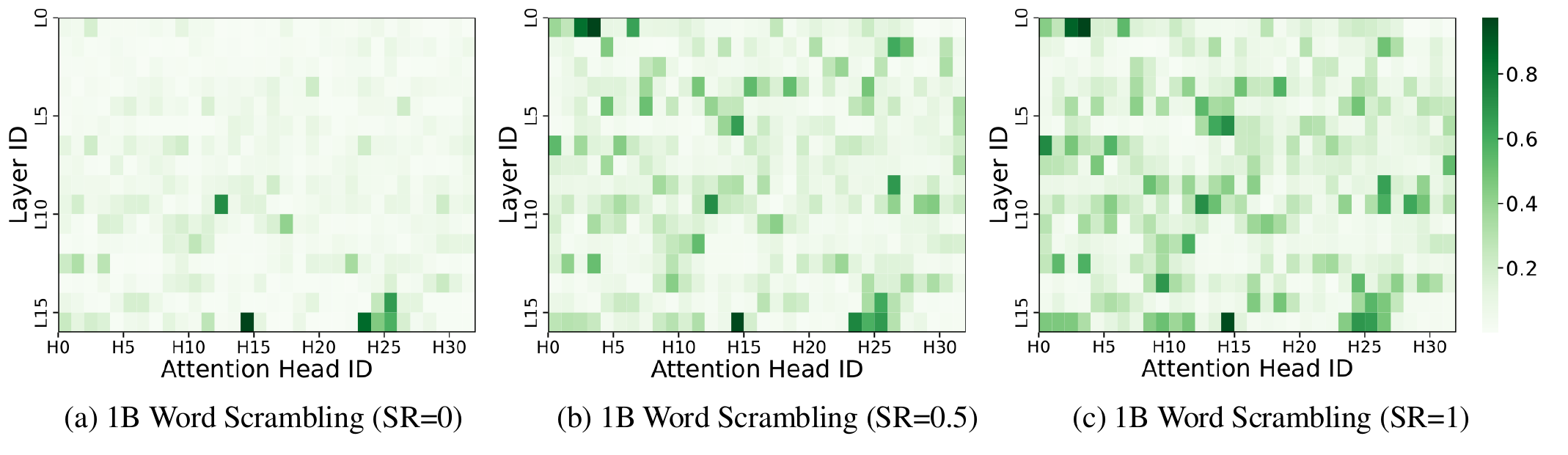}
    \caption{Heatmaps of attention allocation to word form in the LLaMA-1B-Instruct across Scramble Ratios (SR). The x-axis denotes attention heads, and the y-axis denotes layers. \textit{Specific heads consistently focus on word form}, with higher SR activating more form-sensitive heads, indicating a structured and stable processing mechanism.}
    \label{fig:p3_2}
\end{figure*}

\subsection{Attention Allocation to Word Form}\label{sec:attention_allocation}

Since context integrity (CI) has minimal impact on semantic reconstruction, we mainly analyze the model’s attention patterns under CI=1, where the surrounding context remains intact, which aligns with a realistic setting for typoglycemia.

\paragraph{Definition of AttentionSelf.}
To quantify how LLMs allocate attention to word form, we define \textit{AttentionSelf}, which measures the total attention assigned to all tokens in a subword sequence by its final token, aggregated across all attention heads. Formally, we define AttentionSelf as:
\[
\text{AttentionSelf} = \textstyle \sum_{h \in H} \sum_{t \in T} A_h(t_{\text{last}}, t),
\]
where \( H \) represents all attention heads, \( T \) is the set of tokens within the scrambled subword sequence, and \( A_h(t_{\text{last}}, t) \) denotes the attention weight assigned by head \( h \) from the final token \( t_{\text{last}} \) to token \( t \) in the sequence.

We compute the mean AttentionSelf across a set of samples at each layer, as shown in Figures~\ref{fig:p3_1}.

\paragraph{Attention Allocation Pattern.}
Across all models, we observe a consistent trend in which AttentionSelf increases with SR across all layers, with the SR=1 curve consistently the highest and SR=0 significantly lower than the rest. This ordering remains consistent across all three models, with curves ranked from top to bottom according to SR, indicating that higher scrambling severity leads to stronger attention to word form.

Notably, even from Layer 0, LLMs allocate substantial attention to scrambled words (SR > 0), suggesting that word reconstruction begins at the earliest processing stages. This finding aligns with Section~\ref{sec:exp2}, where semantic reconstruction was observed to start from the lowest layers. In contrast, when SR=0, the model assigns minimal attention to word form at the lower layers, with AttentionSelf remaining consistently lower in the initial processing stages. This suggests that LLMs do not explicitly attend to word form in the early layers unless perturbation occurs. 

Across all three models, AttentionSelf exhibits a clear upward trend in the higher layers, indicating that LLMs increasingly refocus on word form at deeper layers. In the 1B and 3B models, we observe a cyclic pattern in attention allocation, suggesting that attention to word form fluctuates across layers rather than following a strictly monotonic trend. In the 70B model, while a similar cyclic pattern appears in the first half of the network, the latter half maintains a prolonged period of low AttentionSelf before a final sharp increase at the highest layers. 

Notably, this final surge in AttentionSelf does not correspond to a continued rise in SemRecScore, which instead plateaus before slightly declining in the last two layers. This behavior is unique to the 70B model; in contrast, the 1B and 3B models exhibit a late-stage increase in both AttentionSelf and SemRecScore, indicating continued refinement of semantic reconstruction. The prolonged low AttentionSelf in the second half of the 70B model aligns with the plateau in SemRecScore observed in Section~\ref{sec:exp2}, suggesting that the model deprioritizes word form processing for a substantial depth range before ultimately refocusing on it in the final layers. Rather than continuously refining the reconstructed semantics, the 70B model appears to reallocate processing resources toward other representational objectives, leading to a deviation from the trends observed in smaller models.

Additionally, we analyze AttentionSelf when CI=0, where all contextual information is removed for the 3B model. 
As shown in Figure~\ref{fig:p3_1}, even with no surrounding context, the model’s attention to word form remains structured and consistent with the pattern observed in Figure~\ref{fig:p3_1}(b), where full context is present. This suggests that LLMs’ attention allocation to word form is a fixed process rather than an adaptive response to available context.

\subsection{Form-Sensitive Attention Heads}\label{sec:form_sensitive_heads}

To further analyze how LLMs utilize word form information, we examine the attention allocation of individual attention heads across all layers and samples. As shown in Figure~\ref{fig:p3_2}, the x-axis represents the \textit{Attention Head ID}, while the y-axis denotes the \textit{Layer ID}. The color intensity of each cell corresponds to AttentionSelf\(_i\), which quantifies the attention allocated to word form by the attention head \(i\). Figure~\ref{fig:p3_1} presents the LLaMA-3.2-1B-Instruct model’s attention distribution under different SR values, while heatmaps for the 3B and 70B models can be found in Appendix~\ref{heatmap_app}.

In Section\ref{sec:attention_allocation}, we observed that all models exhibit an increase in overall attention to word form at the highest layers. Here, Figure~\ref{fig:p3_2} reveals that, across all SR values, attention heads \textit{H14, H24, and H25} in the final layer consistently focus on word form. This suggests that certain attention heads are specifically responsible for processing word form information at the model’s top layers.

When SR=0, the lower layers show minimal attention to word form, which aligns with our previous observation that LLMs do not explicitly attend to word form in early layers unless perturbation occurs. However, when SR>0, even at the lowest layer, \textit{H2 and H3} consistently allocate attention to word form, indicating that the model begins reconstructing scrambled words from the very first processing stages. The cyclic attention pattern observed in Section~\ref{sec:attention_allocation} appears to be primarily driven by \textit{H12 and H26} in the middle layers, reinforcing the idea that certain attention heads exhibit periodic fluctuations in their focus on word form.

Additionally, heatmaps for the 3B and 70B models (see Appendix~\ref{heatmap_app}) confirm that form-sensitive attention heads are consistently present across model scales. As SR increases, more such attention heads are activated, and their distribution remains stable across different SR levels. LLMs’ utilization of word form is primarily carried out by these specific attention heads rather than being distributed uniformly across the model.

These findings suggest that LLMs possess specialized attention heads dedicated to processing word form, which become increasingly engaged as word scrambling severity increases.

\section{Conclusion and Future Work}

Our findings reveal that LLMs primarily rely on word form for typoglycemia-style semantic reconstruction, with contextual information playing a minimal role. We further demonstrate that attention allocation to word form follows a structured pattern across layers, with cyclic fluctuations and specialized attention heads dedicated to word form processing.
Future work should explore a broader range of model architectures and languages, as well as assess the practical impact of LLMs’ word form utilization on real-world NLP tasks.

\section{Limitations}

While our study offers valuable insights into how LLMs utilize word form for semantic reconstruction, it has several limitations. First, our experiments are confined to the LLaMA model family, leaving open the question of whether these mechanisms generalize to other architectures. Second, we focus specifically on typoglycemia-style scrambling, whereas other perturbations, such as deletions or phonetic errors, may lead to different reconstruction patterns. Finally, our analysis is limited to English, and it remains uncertain whether morphologically rich languages exhibit similar dependencies on word form.
Future work could explore broader model architectures, diverse perturbation types, and cross-linguistic analyses to provide a more comprehensive understanding of LLMs’ semantic reconstruction mechanisms.

\bibliography{custom}
\clearpage
\appendix

\section{Semantic reconstruction performance}

\subsection{Performance across different CI levels}\label{exp2_app1}
\begin{figure*}[tb]
    \centering
    \includegraphics[width=1\linewidth]{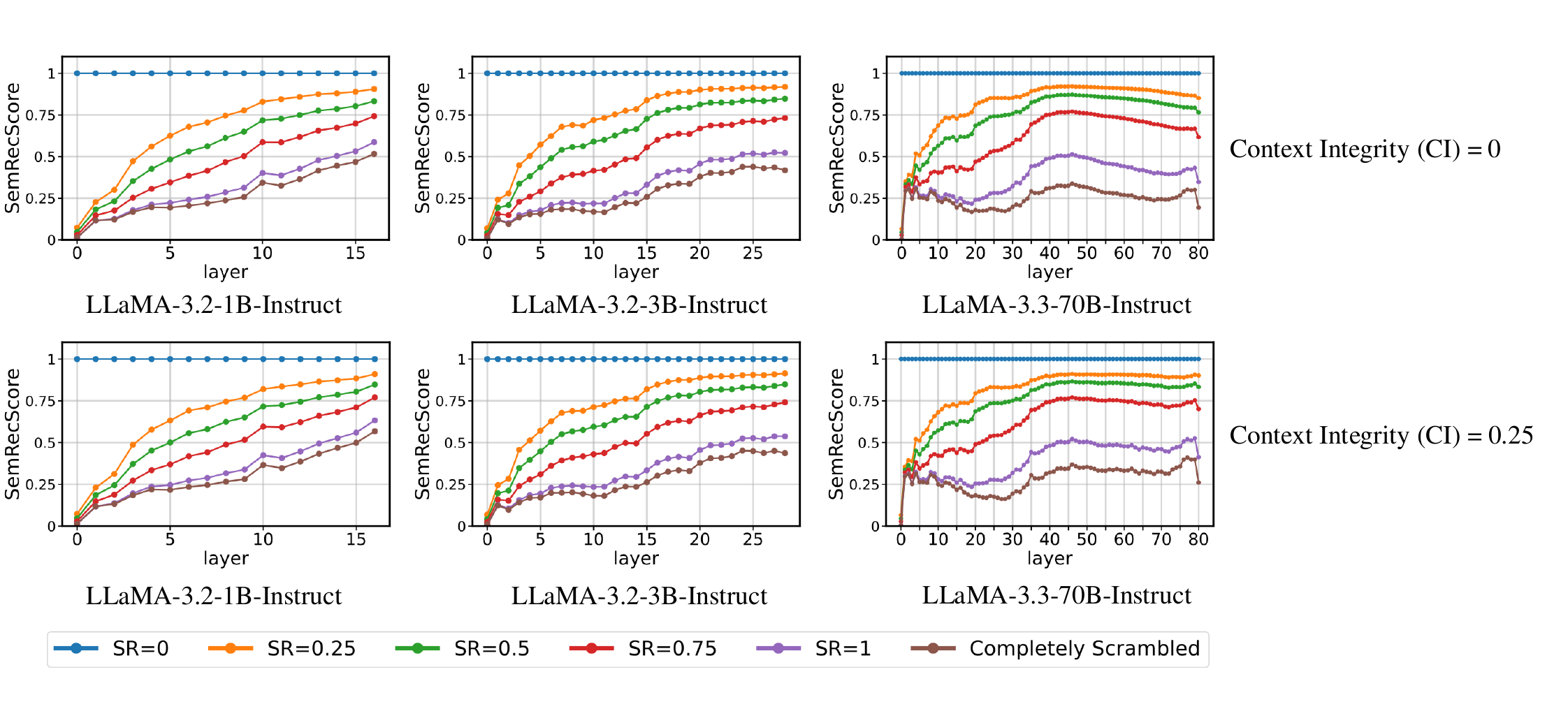}
    \caption{Semantic Reconstruction Performance across Different LLM Scales and Context Integrity Levels.
The plots illustrate the layer-wise Semantic Reconstruction Score (SemRecScore) for various SR values across different LLaMA models (1B, 3B, and 70B). The top row represents CI = 0, while the bottom row represents CI = 0.25. The legend indicates different SR conditions, including the “Completely Scrambled” setting. The similarity of the curves across different CI values suggests that Context Integrity (CI) has minimal impact on semantic reconstruction performance.}
    \label{fig:appendix1}
\end{figure*}

\begin{figure*}[tb]
    \centering
    \includegraphics[width=1\linewidth]{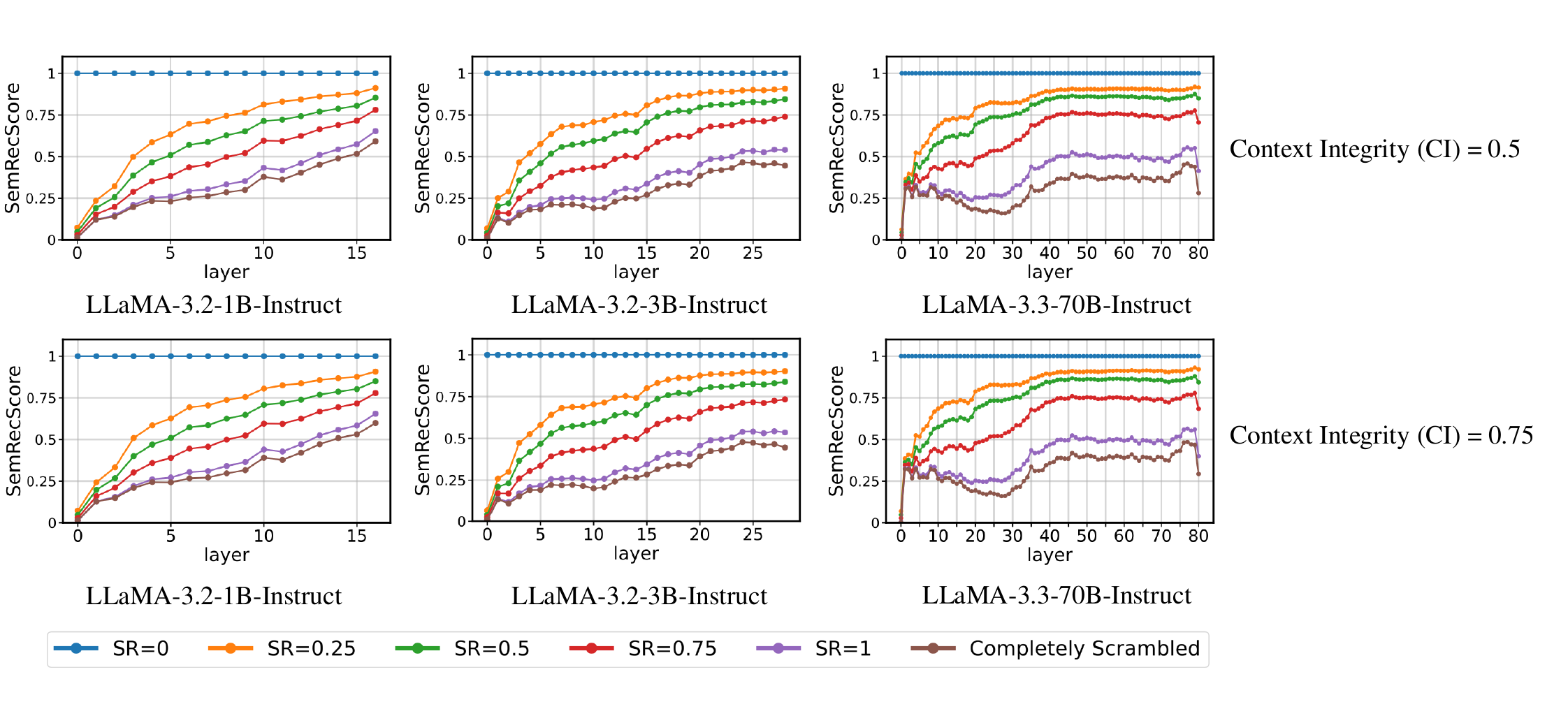}
    \caption{Semantic Reconstruction Performance across Different LLM Scales and Context Integrity Levels.
The plots illustrate the layer-wise Semantic Reconstruction Score (SemRecScore) for various SR values across different LLaMA models (1B, 3B, and 70B). The top row represents CI = 0.25, while the bottom row represents CI = 0.75. The legend indicates different SR conditions, including the “Completely Scrambled” setting. The similarity of the curves across different CI values suggests that Context Integrity (CI) has minimal impact on semantic reconstruction performance.}
    \label{fig:appendix2}
\end{figure*}

In the main text, we have presented the layer-wise Semantic Reconstruction Performance across different LLM scales for various SR values when CI = 1. In this appendix section, Figure~\ref{fig:appendix1} illustrates the results for CI = 0 and CI = 0.25, while Figure~\ref{fig:appendix2} presents the results for CI = 0.5 and CI = 0.75. The similarity of the curves confirms that Contextual Information has a minimal impact on semantic reconstruction.

\subsection{Performance across different SR levels}\label{exp2_app2}

\begin{figure*}[tb]
    \centering
    \includegraphics[width=1\linewidth]{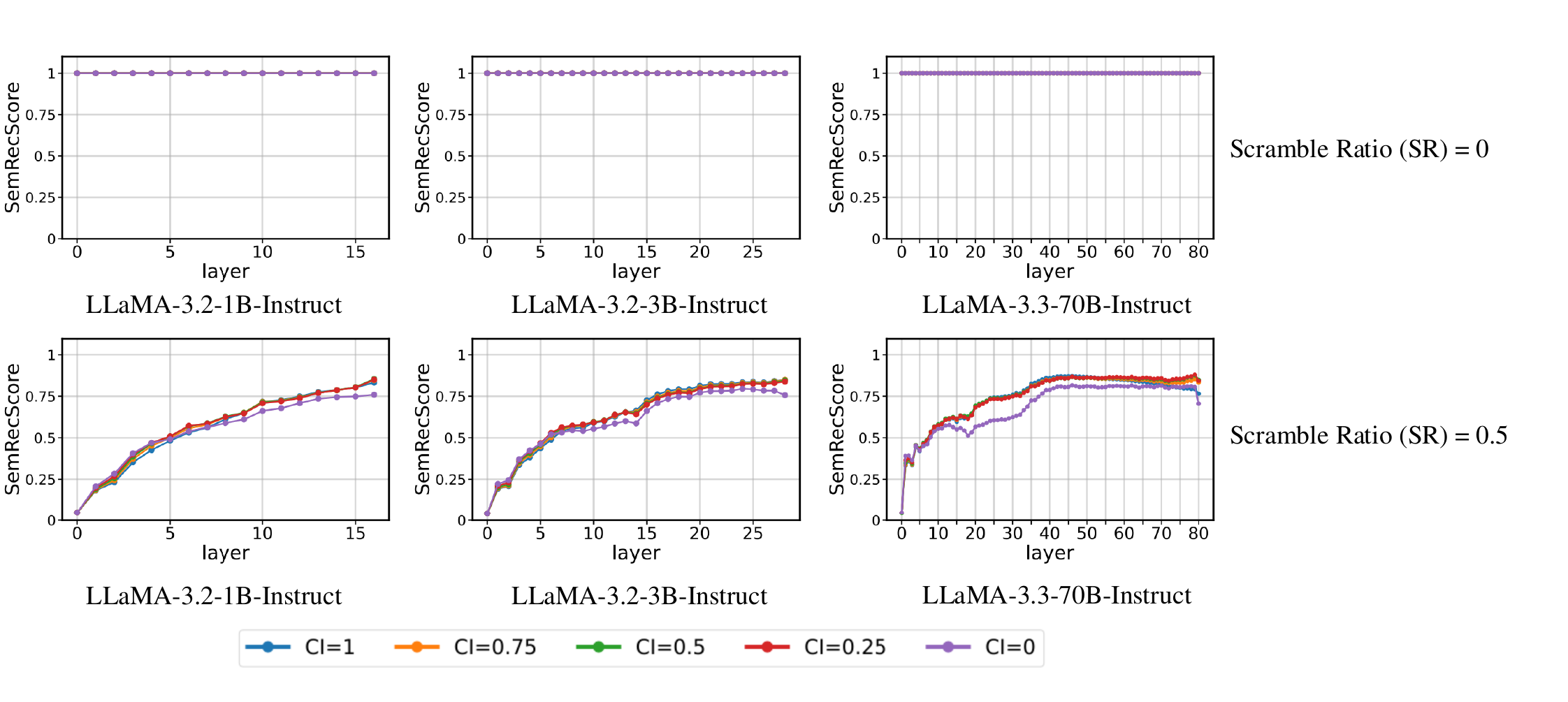}
    \caption{Semantic Reconstruction Performance across Different LLM Scales and Scramble Ratio Levels.
The plots illustrate the layer-wise Semantic Reconstruction Score (SemRecScore) for various CI values across different LLaMA models (1B, 3B, and 70B). The top row represents SR = 0, while the bottom row represents CI = 0.5. The legend indicates different CI conditions.The close alignment of curves across different CI values suggests that Context Integrity has a limited impact on semantic reconstruction.}
    \label{fig:appendix3}
\end{figure*}

\begin{figure*}[tb]
    \centering
    \includegraphics[width=1\linewidth]{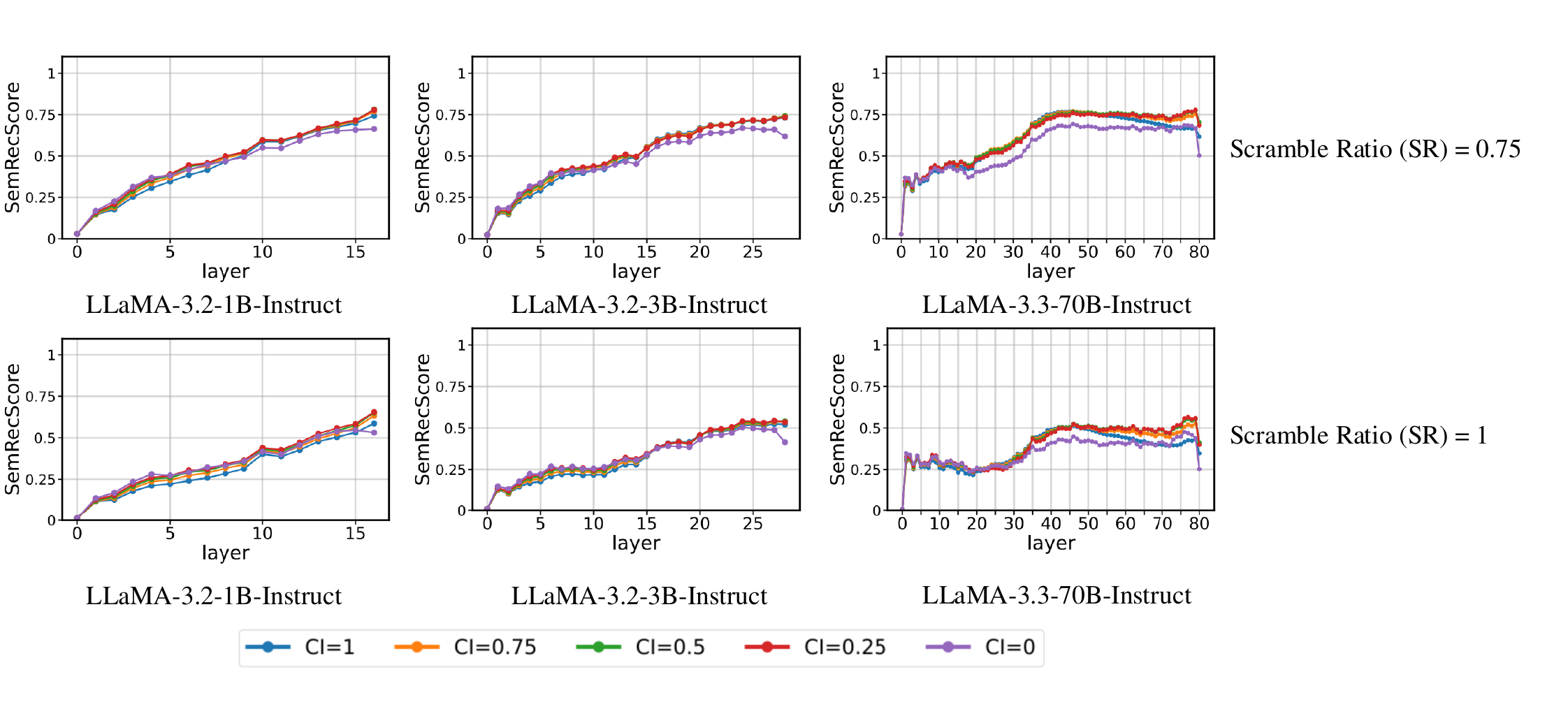}
    \caption{Semantic Reconstruction Performance across Different LLM Scales and Scramble Ratio Levels.
The plots illustrate the layer-wise Semantic Reconstruction Score (SemRecScore) for various CI values across different LLaMA models (1B, 3B, and 70B). The top row represents SR = 0.75, while the bottom row represents CI = 1. The legend indicates different CI conditions.The close alignment of curves across different CI values suggests that Context Integrity has a limited impact on semantic reconstruction. In the rows with higher SR, all curves are noticeably lower, confirming that Word Form plays a crucial role in semantic reconstruction.}
    \label{fig:appendix4}
\end{figure*}

In the main text, we have presented the layer-wise Semantic Reconstruction Performance across different LLM scales for various CI values when SR = 0.25. In this appendix section, Figure~\ref{fig:appendix3} illustrates the results for SR = 0 and SR = 0.5, while Figure~\ref{fig:appendix4} presents the results for SR = 0.75 and SR = 1. The noticeable decline in the curves as SR increases confirms that Word Form plays a dominant role in semantic reconstruction.

\section{Heatmap of attention allocaton}\label{heatmap_app}

In the main text, we presented how different attention heads in LLaMA-3.2-1B-Instruct allocate attention to word forms. This section provides heatmaps of attention distribution for the remaining two models.

\begin{figure*}[tb]
    \centering
    \includegraphics[width=1\linewidth]{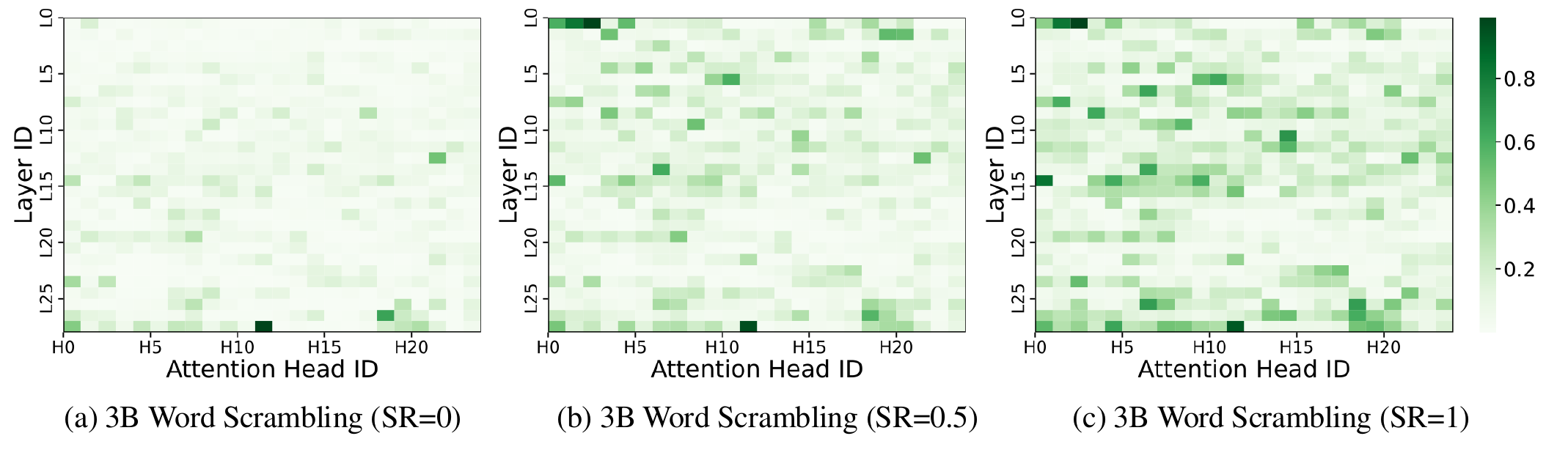}
    \caption{Heatmaps of attention allocation to word form in the LLaMA-3.2-3B-Instruct across Scramble Ratios (SR). The x-axis denotes attention heads, and the y-axis denotes layers. \textit{Specific heads consistently focus on word form}, with higher SR activating more form-sensitive heads, indicating a structured and stable processing mechanism.}
    \label{fig:appendix5}
\end{figure*}

\begin{figure*}[tb]
    \centering
    \includegraphics[width=1\linewidth]{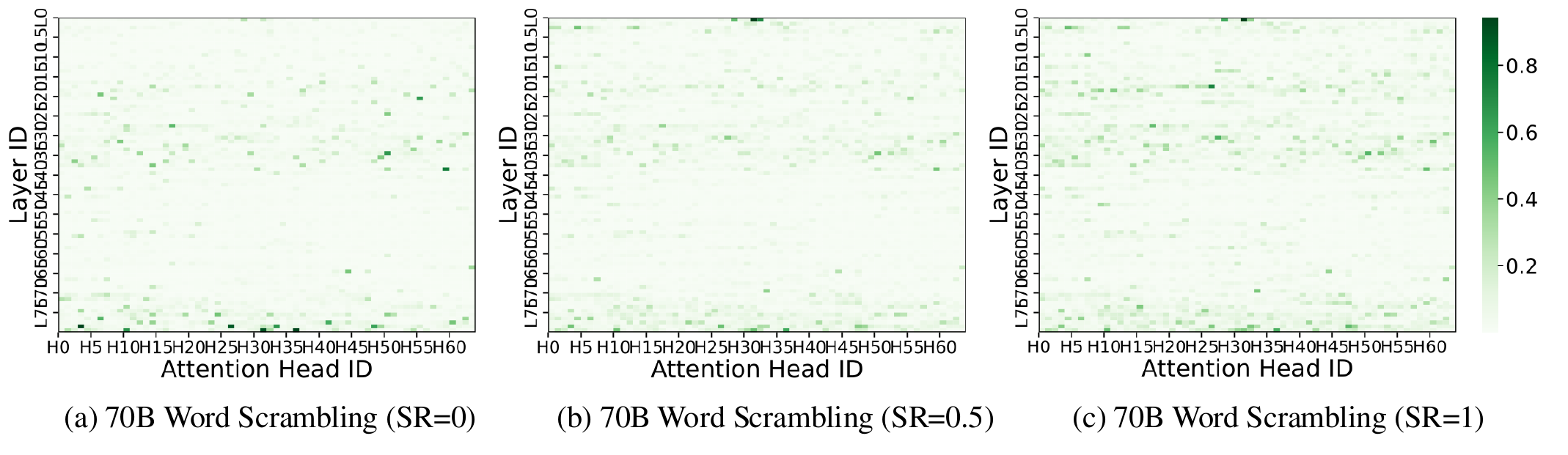}
    \caption{Heatmaps of attention allocation to word form in the LLaMA-3.3-70B-Instruct across Scramble Ratios (SR). The x-axis denotes attention heads, and the y-axis denotes layers. \textit{Specific heads consistently focus on word form}, with higher SR activating more form-sensitive heads, indicating a structured and stable processing mechanism.}
    \label{fig:appendix6}
\end{figure*}

Figure~\ref{fig:appendix5} presents the attention distribution heatmap for LLaMA-3.2-3B-Instruct, while Figure~\ref{fig:appendix6} shows the heatmap for LLaMA-3.3-7B-Instruct. Across different SR levels, a consistent pattern emerges: certain layers and specific attention heads consistently allocate more attention to word forms. This observation suggests that the models primarily rely on Form-Sensitive Attention Heads to utilize word form information.

\end{document}